\documentclass[conference]{IEEEtran}
\IEEEoverridecommandlockouts

\usepackage{cite}
\usepackage{amsmath,amssymb,amsfonts}
\usepackage{algorithmic}
\usepackage{graphicx}
\usepackage{textcomp}
\usepackage{xcolor}
\usepackage{kotex}
\usepackage{tabulary}
\usepackage{multirow}
\usepackage{comment}
\usepackage{url}

\def\BibTeX{{\rm B\kern-.05em{\sc i\kern-.025em b}\kern-.08em
    T\kern-.1667em\lower.7ex\hbox{E}\kern-.125emX}}

\begin{document} 

\title{An Empirical Experiment on Deep Learning Models for Predicting Traffic Data}
\author{
  \IEEEauthorblockN{
    Hyunwook Lee\IEEEauthorrefmark{2}\quad
    Cheonbok Park\IEEEauthorrefmark{3}\quad
    Seungmin Jin\IEEEauthorrefmark{2}\quad
    Hyeshin Chu\IEEEauthorrefmark{2}\quad
    Jaegul Choo\IEEEauthorrefmark{1}\IEEEauthorrefmark{4}\quad
    Sungahn Ko\IEEEauthorrefmark{1}\IEEEauthorrefmark{2}\thanks{\IEEEauthorrefmark{1}Corresponding author}
  }
  \vspace{1mm}
  \IEEEauthorblockA{
    \IEEEauthorrefmark{2}Ulsan National Institute of Science and Technology, 
    \IEEEauthorrefmark{3} NAVER,
    \IEEEauthorrefmark{4}Korea Advanced Institute of Science and Technology
    \vspace{1mm}\\
    \{gusdnr0916, skyjin, hyeshinchu, sako\}@unist.ac.kr, cbok.park@navercorp.com, jchoo@kaist.ac.kr
  }
}

\maketitle

\begin{abstract}
To tackle ever-increasing city traffic congestion problems, researchers have proposed deep learning models to aid decision-makers in the traffic control domain. 
Although the proposed models have been remarkably improved in recent years, there are still questions that need to be answered before deploying models. 
For example, it is difficult to figure out which models provide state-of-the-art performance, as recently proposed models have often been evaluated with different datasets and experiment environments. 
It is also difficult to determine which models would work when traffic conditions change abruptly (e.g., rush hour). 
In this work, we conduct two experiments to answer the two questions. 
In the first experiment, we conduct an experiment with the state-of-the-art models and the identical public datasets to compare model performance under a consistent experiment environment. 
We then extract a set of temporal regions in the datasets, whose speeds change abruptly and use these regions to explore model performance with difficult intervals.
The experiment results indicate that Graph-WaveNet and GMAN show better performance in general. 
We also find that prediction models tend to have varying performances with data and intervals, which calls for in-depth analysis of models on difficult intervals for real-world deployment. 

\end{abstract}
\begin{IEEEkeywords}
Deep Learning, Traffic Forecasting, Speed Prediction, Flow Prediction
\end{IEEEkeywords}

\section{Introduction}
As traffic problems have become severe in major cities, a considerable amount of research has been performed to forecast traffic conditions, such as traffic speed and flow, utilizing machine learning~\cite{Vlahogianni14, Li18}.
Recently, as deep learning models have become dominant in various domains, novel deep learning models to solve traffic data prediction problems have also increased~\cite{Xie20, Yin20}. 
Although approaches have been proposed, some issues remain unsolved in traffic prediction. 
First, as previous models are evaluated with different datasets and under different experimental environments~\cite{Li18}, comparing the accuracy and computation times of diverse models is challenging.
Second, numerous existing studies utilize average accuracy to evaluate models, but this approach is not conducive to performing a thorough analysis. 
For example, both recurring (e.g., daily and rush hour) and non-recurring (e.g., accident) patterns exist, and some of them change abruptly (e.g., sudden decrease in speed due to accidents) in the speed data~\cite{Sun17}. 
In many cases, it is more critical to suggest accurate predictions for non-flat and abruptly changing speed patterns for real-world use cases~\cite{Lee20, Li18}. 

In this work, we conduct a series of experiments to resolve the aforementioned issues. 
First, we review and describe different traffic speed and flow datasets. 
Then, we select eight models and seven datasets to directly compare the accuracy and computation times across the models and datasets in the same running environment. 
The selected models are STGCN~\cite{yu18stgcn}, DCRNN~\cite{li18dcrnn}, ASTGCN~\cite{guo19astgcn}, ST-MetaNet~\cite{pan19metanet}, Graph-WaveNet~\cite{wu19gwnet}, STG2Seq~\cite{bai19stg2seq}, STSGCN~\cite{song20stsgcn}, and GMAN~\cite{zheng20gman}. 
We include METR-LA, PeMS-BAY, and PeMSD7 as speed datasets and PeMSD3, PeMSD4, PeMSD7, and PeMSD8 as flow datasets.
We also extract intervals during which traffic conditions change abruptly.

The experimental results indicate that Graph-WaveNet~\cite{wu19gwnet} is generally the most accurate across datasets. However, GMAN~\cite{zheng20gman} performed the best when making long-term predictions, such as 60-minutes prediction.
In terms of computation time, STGCN requires the least amount of training time, while Graph-WaveNet is the fastest in producing prediction results. 
In an experiment involving intervals with abruptly changing speed conditions, we obtain similar results, as Graph-WaveNet has the highest accuracy rates, and GMAN's performance is superior in long-term prediction.

Several papers similar to our work have been published. 
For example, Vlahogianni et al. and Li et al. conduct surveys on traffic prediction models, focusing on short-term prediction~\cite{Vlahogianni14, Li18, Yin20}. 
Yin et al.~\cite{Yin20} also present a survey on traffic prediction from multiple perspectives, including methods, applications, datasets, and experiments.
Our work differs from prior work, as we report and compare the accuracy and computation time of state-of-the-art prediction models with multiple speed and flow datasets in the same environment.

\begin{table*}[t]
\caption{Summary of PeMS datasets used for predicting traffic speed and flow. }
\centering
\resizebox{1.9\columnwidth}{!}{\begin{tabular}{c|c|c|c|c|c|c|c}
\hline
        Tasks         & \multicolumn{3}{c|}{Speed prediction} & \multicolumn{4}{c}{Flow Prediction} \\
\hline
\hline
        Name          & METR-LA     & PeMS-BAY  & PeMSD7(M)         & PeMSD3        & PeMSD4                 & PeMSD7            & PeMSD8\\
        Region        & Los Angeles & Bay Area  & Los Angeles       & North Central & Bay Area               & Los Angeles       & San Bernardino\\
        Start Date    & 3/1/2012    & 1/1/2017  & 5/1/2012          & 9/1/2018      & 1/1/2018               & 5/1/2017          & 7/1/2016\\
        End Date      & 6/30/2012   & 6/30/2017 & 6/30/2012         & 11/30/2018    & 2/28/2018              & 8/31/2017         & 8/31/2016\\
        \# of Days       & 122         & 181       & 44$^{\mathrm{a}}$ & 91            & 59                     & 98$^{\mathrm{b}}$ & 62\\
        \# Nodes      & 207         & 325       & 228               & 358           & 307                    & 883               & 170\\
        Features      & speed       & speed     & speed             & flow          & flow, occupancy, speed & flow              & flow, occupancy, speed\\
        Sensor ID     & Y           & Y         & N                 & Y             & N                      & N                 & N\\
        citation      & \cite{li18dcrnn, chen19, wu19gwnet, pan19metanet, zhang18gaan, wang18, chen20}  & \cite{li18dcrnn, chen19, wu19gwnet, zheng20gman, chen20}  & \cite{yu18stgcn, yu19}  & \cite{song20stsgcn}              &  \cite{shi20aptn, guo19astgcn, guo20, song20stsgcn}                      &     \cite{song20stsgcn}              & \cite{shi20aptn, guo19astgcn, song20stsgcn}\\
\hline
        \multicolumn{8}{l}{$^{\mathrm{a}}$Only contains weekdays}\\
        \multicolumn{8}{l}{$^{\mathrm{b}}$Real data range and one in the original paper is not equivalent}
\end{tabular}}
\label{tbl_data_desc}
\vspace{-4mm}
\end{table*}
The main contributions of this work include the following: 
\begin{itemize}
    \item Experiments to compare the accuracy and time consumption of eight state-of-the-art deep learning models using seven different speed and flow datasets;
    \item A report of the accuracy of the selected models with intervals of abruptly changing traffic condition, and 
    \item Lessons learned from the experiments.
\end{itemize}

\section{Related Work}
In this section we describe existing surveys on deep learning models for traffic data prediction. 
Vlahogianni et al.~\cite{Vlahogianni14} review conventional approaches for predicting short-term traffic data in the categories of 10 challenges, including data resolution, aggregation and quality, and fusing data from other sources.
Reviewing traditional methods used for traffic prediction is helpful to understanding existing methods. However, the methods do not cover deep learning approaches, which are drawing a considerable amount of attention in the traffic prediction literature.
Furthermore, they do not discuss model performance and datasets.
Li et al.~\cite{Li18} categorize prior work based on whether spatial modeling methods are used. 
In addition, they explain existing research problems and challenges. 
For example, they describe traffic prediction in extreme conditions, such as incidents, as an important challenge.
However, they do not consider any datasets nor make performance comparisons.
Xie et al.~\cite{Xie20} focus on existing models that use flow datasets. 
They describe data processing methods and significant factors affecting accurate flow prediction, as well as categorize existing methods as based on statistics, traditional machine learning, deep learning, reinforcement learning, or transfer learning.
However, their work does not cover model evaluation. 
Yin et al.~\cite{Yin20} present a comprehensive survey on traffic prediction tasks in multiple perspectives, including methods, applications, datasets, and experiments. 
They provide a taxonomy for existing works, summarize public datasets for forecasting tasks, and compare the performance of the models. 
However, the existing work utilizes only a few datasets in different running environments, which causes difficulty in finding superiority of a particular models in the surveys.

In this work, we perform a series of extensive experiments with eight state-of-the-art models in the same environment to make direct comparisons of the models across seven different speed and flow datasets. 
We also extract difficult intervals and perform an experiment with the intervals to reveal which models perform better than others during intervals with abruptly changing conditions.

\section{Data Description}
In this section, we describe Performance Measurement System (PeMS) datasets~\cite{chao00, pems}, which have been used extensively for speed and flow prediction tasks.
PeMS is a system operated by the California Department of Transportation (Caltrans)~\cite{pems}, which collects traffic data from more than 45,000 independent detectors installed across freeways in major metropolitan areas of California. 
It collects data from each detector every 30 seconds and aggregates them into five-minute interval values by lane. 
It also provides distance information of roads which can be used to build a road network graph and its adjacency matrix. 

Most of the existing approaches that use the PeMS data for speed and flow prediction~\cite{yu18stgcn, li18dcrnn, chen19, wu19gwnet, pan19metanet, zhang18gaan, zheng20gman, shi20aptn, guo19astgcn, wang18, chen20, guo20, song20stsgcn, yu19} utilize the five-minute aggregated values.
We summarize the data used in the existing work in Table~\ref{tbl_data_desc}, using eight categories-- dataset name, the region, start and end dates of the data, the data range, sensor counts, features used, and sensor IDs. 
To supplement existing datasets for further experiments, we present the sensor IDs, which can be used to access data for specific IDs in PeMS. 
For example, while the PeMSD3 dataset does not contain speed data, one can download it for each sensor and data range from the PeMS site using sensor IDs in the data.
\begin{table*}[t]
    \centering
    \caption{Characterization of spatial and temporal modeling methods}
    \resizebox{1.9\columnwidth}{!}{\begin{tabular}{c|c|c|c|c}
        \hline
        & Component & Pros & Cons & Models\\
        \hline\hline
        \multirow{5}{*}{Spatial} & GCN~\cite{Bruna14} & \begin{tabular}{c}
            Simple Architecture  \\
            Direct use of graph structures
        \end{tabular} & \begin{tabular}{c}
            K-hop neighboring problem\\
            Cannot consider a graph structure change
        \end{tabular} & \begin{tabular}{c}
            STGCN$^1$, DCRNN$^2$, ASTGCN$^1$, \\
            Graph-WaveNet$^2$, STG2Seq$^2$, STSGCN$^2$
        \end{tabular}\\
        \cline{2-5}
        & GAT~\cite{Velickovic17} & \begin{tabular}{c}
            Dynamic modeling of spatial correlation\\
            Interpretability
        \end{tabular}
        & \begin{tabular}{c}
            High time and memory cost\\
        \end{tabular} & \begin{tabular}{c}
            ST-MetaNet
        \end{tabular}\\
        \cline{2-5}
        & Attn+Graph Embedding~\cite{zheng20gman} & \begin{tabular}{c}
            Dynamic modeling of spatial correlation\\
            Consideration of latent features\\
            Attention beyond the graph structure
        \end{tabular}
        & \begin{tabular}{c}
            Random grouping corrupts graph structures\\
        \end{tabular} & \begin{tabular}{c}
            GMAN
        \end{tabular}\\
        \hline
        \multirow{5}{*}{Temporal} &  RNN~\cite{Siegelmann91} & \begin{tabular}{c}
            Consideration of all states 
        \end{tabular}
        & \begin{tabular}{c}
            Complex architecture\\
            Hard to capture local hidden feature
        \end{tabular} & \begin{tabular}{c}
            DCRNN, ST-MetaNet
        \end{tabular}\\
        \cline{2-5}
        & CNN~\cite{Lecun10cnn}  & \begin{tabular}{c}
            Simple architecture\\
            Good at local feature extraction\\
            Prediction for multiple steps at once\\
        \end{tabular}
        & \begin{tabular}{c}
            Should find the best filter size
        \end{tabular} & \begin{tabular}{c}
            STGCN, ASTGCN, Graph-WaveNet\\
        \end{tabular}\\
        \cline{2-5}
        & Attention~\cite{Vaswani17}  & \begin{tabular}{c}
            Flexible feature selection\\
            Less time consumption on referring long-term data
        \end{tabular}
        & \begin{tabular}{c}
            Generally high time/memory cost
        \end{tabular} & \begin{tabular}{c}
            ASTGCN, GMAN\\
        \end{tabular}\\
        \hline
        \multicolumn{3}{l}{$^1$spectral-based GCN, $^2$spatial-based GCN}
    \end{tabular}}
    \label{tbl_component}
    \vspace{-4mm}
\end{table*}

We find that METR-LA and PeMS-BAY data are the most popular for speed prediction.
Originally, there are 4,573 and 3,656 sensors in the Los Angeles and Bay area, respectively, but only a sampled sensors in the region is used for training (207 and 305 sensors for 122 and 181 days in the METR-LA and PeMS-BAY datasets, respectively).
Compared to PeMSD7(M), they contain both weekday and weekend data. 
PeMSD4 and PeMSD8 have been frequently used to predict traffic flow. 
They include data for about two months and provide not only flow information, but also occupancy and speed. 
Note that the data described in this section is not complete, since we exclude taxi- and bike-demand datasets and focus on speed and flow data, which can be represented with graphs.
For a more comprehensive dataset collection, we refer readers to the survey presented by Yin et al.~\cite{Yin20}.
Next, we will describe how we select the models and present experiments to compare their performance.

\section{Preliminaries}
\subsection{Model Selection Process}
Deep learning models for traffic data aim to effectively model the spatial and temporal dependencies of roads, utilizing a graph structure of road networks~\cite{Li18}. 
To summarize existing models for traffic prediction and select the models for the experiments, we take several steps. 
First, we review existing surveys on traffic prediction (e.g., \cite{Yin20}). 
Then, we individually explore prestigious venues for publications, including AAAI, IJCAI, and ITS. 
Next, we further search for publications in the IEEE Xplore and ACM digital libraries.
Finally, we find 35 papers from our search.
Prior to characterization of the different models (Table~\ref{tbl_component}), 
we exclude 20 models that do not consider graph structures for spatial modeling, which results in lower accuracy when compared to other models. 
We also exclude the papers whose source code and datasets are not publicly accessible.
Finally, we choose eight deep learning models to predict traffic speed and flow. 

Table~\ref{tbl_component} summarizes the spatial and temporal components and pros and cons of the chosen models.
We note that there are two types of graph convolutional networks (GCNs)--spatial-based and spectral-based.
Spatial-based GCNs apply convolution directly to the adjacency matrix, utilizing physical distances and connections among roads. 
In contrast, the spectral-based GCNs use a Laplacian matrix, a graph structure representation in the spectral domain, for graph convolution. 

\subsection{Problem Statement}
We begin by defining the road network structure and its representation, followed by reviewing and summarizing the problem statements of existing work. 
To predict traffic data, we define the road network graph as $\mathcal{G} = (\mathcal{V}, \mathcal{E}, \mathcal{A})$, where $\mathcal{V}$ is the set of vertices (i.e., sensors) with $|\mathcal{V}|=N$, $\mathcal{E}$ as the set of the edges, representing the connectivity between roads, and $\mathcal{A}\in \mathbb{R}^{N\times N}$ is a weighted adjacency matrix that contains the connectivity and edge weight information. 
Edge weights are calculated based on the distance and direction of the edges between two connected nodes. 
If $\mathcal{A}$ is a simple non-weighted adjacency matrix, then the edge weights are either 0 or 1, and $\mathcal{A}$ only represents connectivity. 
If $\mathcal{A}$ is a weighted adjacency matrix, most of the previous approaches calculate edge weights with a Gaussian kernel, as follows: $W_{ij} = \exp{-\frac{{\text{dist}}_{ij}^2}{\sigma^2}}$~\cite{li18dcrnn, wu19gwnet, yu18stgcn}, where dist$_{ij}$ is a distance between sensor $i$ and $j$ and $\sigma$ is the standard deviation of the distances.

Note that a road network can be directly represented by the weighted adjacency matrix $\mathcal{A}$.
Traffic forecasting is a typical spatiotemporal data prediction problem, which aims to predict a value in the next $T$ time steps with previous $T'$ historical traffic data and an adjacency matrix.
Traffic data at time $t$ is represented by a graph signal matrix $X_\mathcal{G}^t \in \mathbb{R}^{N \times C}$, where $C$ is the number of features (e.g., speed, flow, and the normalized time of the day). 
In summary, the deep learning-based traffic prediction task is to learn a mapping function $f(\cdot)$ that predicts future $T$ graph signals from $T'$ historical input graph signals:
\begin{align*}
    \big[{X_\mathcal{G}}^{(t-T'+1)}, \cdots, {X_\mathcal{G}}^{(t)}\big] &\xrightarrow{f(\cdot)} \big[{X_\mathcal{G}}^{(t+1)}, \cdots, {X_\mathcal{G}}^{(t+T)}\big]
\end{align*}

\section{Experiments}
In this section, we describe how we present experimental results. 
We use a server equipped with an Intel Xeon 5120 CPU, 394 GB of RAM, and eight Nvidia Titan RTX GPUs for all the experiments. 
To evaluate both computation time and accuracy of the models, we use Mean Absolute Error (MAE), Root Mean Squared Error (RMSE), and Mean Absolute Percentage Error (MAPE) as evaluation metrics and set the batch size as 64. 
All the datasets are divided into train, validate, and test sets at a 7:1:2 ratio, respectively. 
We implement all models with PyTorch\footnote{\url{https://pytorch.org/}} to avoid any performance differences due to the tools and their settings.
To implement GAT, we use Deep Graph Library (DGL)~\cite{Wang19}, a state-of-the-art library for graph modeling.
We set both $T$ and $T'$ as 12 for fairness. 
We pre-process the data to have two input features--time stamp and speed (or flow). 
Then, we use z-score normalization for the traffic data and min-max normalization for the timestamp.
Lastly, we utilize the same hyperparameter settings from the original work. 
We repeat each experiment five times and use average and standard deviation values to present the results.

\begin{figure*}[t]
    \centering
    \includegraphics[width=1.9\columnwidth]{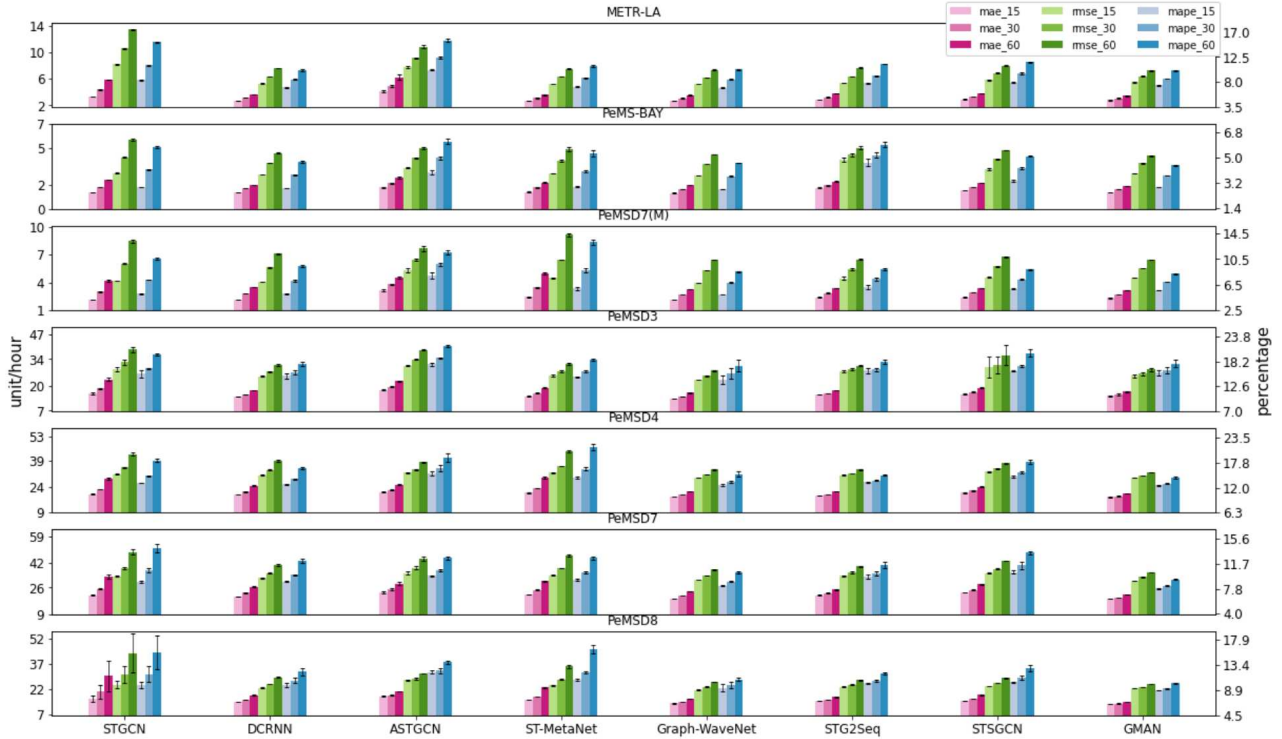}
    \caption{Model comparison results with standard deviation, measured by MAE, RMSE, and MAPE. The above 3 plots for speed data, and the others are plots for flow data.}
    \label{fig:Model_Evaluation}
    \vspace{-4mm}
\end{figure*}
\begin{table}[t]
\caption{Computation time of the models with the METR-LA dataset}
\centering
\resizebox{0.95\columnwidth}{!}{
\begin{tabular}{l|c|c|c}
\hline
                & Training time/epoch  & Inference time     & \# of params  \\
\hline\hline
STGCN           & \textbf{14.8 secs}   & 16.70 secs         & 320k          \\
DCRNN           & 122.22 secs          & 13.44 secs         & 372k          \\
ASTGCN          & 92.11 secs           & 13.68 secs         & 721k          \\
ST-MetaNet      & 135.32 secs          & 24.10 secs         & \textbf{85k}  \\
Graph-WaveNet   & 48.07 secs           & \textbf{3.69 secs} & 309k          \\
STG2Seq         & 87.74 secs           & 11.84 secs         & 351k          \\
STSGCN          & 110.44 secs          & 11.34 secs         & 1100k         \\
GMAN            & 312.1 secs           & 33.7 secs          & 901k          \\
\hline
\end{tabular}}
\label{table:inference}
\vspace{-4mm}
\end{table}

\subsection{Model Performance Experiment Results}
In this section, we present the models' accuracy and computation time for speed and flow data in 15-, 30-, and 60-minute intervals, respectively.
Fig.~\ref{fig:Model_Evaluation} shows the experimental results for speed and flow.

First, we note that Graph-WaveNet outperforms other models for the 15 and 30 minutes interval predictions across all the speed datasets.
For the long-term prediction (i.e., 60-minute interval), GMAN records higher accuracy than other models in general, followed by DCRNN and ST-MetaNet. ASTGCN tends to have the lowest accuracy across all speed datasets.

We can observe a similar result in flow prediction.
Graph-WaveNet and GMAN show higher accuracy than other models in general. 
We also find that GMAN's strength lies in the long-term prediction.
STG2Seq shows the third most accurate prediction, while STGCN records the lowest performance in general.
With the flow datasets, we observe that all models perform better with PeMSD3 and PeMSD8 in terms of MAE and RMSE.
Lastly, Graph-WaveNet performs better with the PeMSD3 and PeMSD8 datasets, while GMAN shows higher accuracy with PeMSD4 and PeMSD7.

When considering both speed and flow prediction, we find that models with spatial-based GCNs (DCRNN, Graph-WaveNet, STSGCN, and STG2Seq) have higher accuracy than those with spectral-based GCNs (STGCN and ASTGCN) in general. 
We also find that including an attention mechanism in temporal modeling is more effective than other methods in long-term prediction.
On the other hand, we notice that STGCN has the highest performance drop across the intervals.
This result shows a weakness of the many-to-one model, which is trained to predict only one time step.
When we compare RNN-based models (DCRNN and ST-MetaNet) to other models, we find that they suffer from error accumulation, which shows the inherent problem of sequence-to-sequence structures. 

\begin{figure*}[t]
    \centering
    \includegraphics[width=1.95\columnwidth]{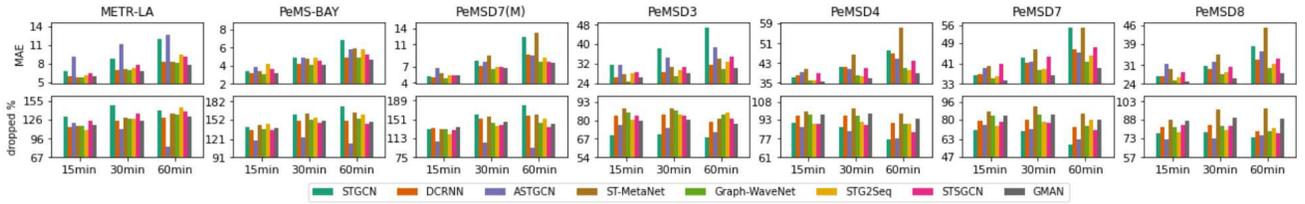}
    \caption{MAE and relative performance degradation in percentage in abruptly changing condition with METR-LA dataset}
    \label{table:impeded}
\end{figure*}

Table~\ref{table:inference} indicates the computation time spent on prediction. 
From the table, we observe that models using GCNs for spatial dependency (e.g., STGCN, DCRNN, ASTGCN, Graph-WaveNet, STG2Seq, and STSGCN) spend less time during both training and inference. 
STSGCN requires the largest number of parameters, as it uses individual modules to capture heterogeneity of the traffic data.
STGCN requires the shortest training time per epoch, but it needs a longer inference time because it has a many-to-one architecture that needs to predict multiple steps separately.

\subsection{Model Performance with Difficult Intervals}
All the existing studies on the traffic prediction models evaluate performance using the average accuracy. 
However, such an evaluation method is insufficient to reveal models' weaknesses and characteristics. 
For example, it is possible that a model could record higher performance than others because it records higher accuracy than others when traffic conditions are stable. 
\begin{figure}[t]
    \centering
    \includegraphics[width=.95\columnwidth]{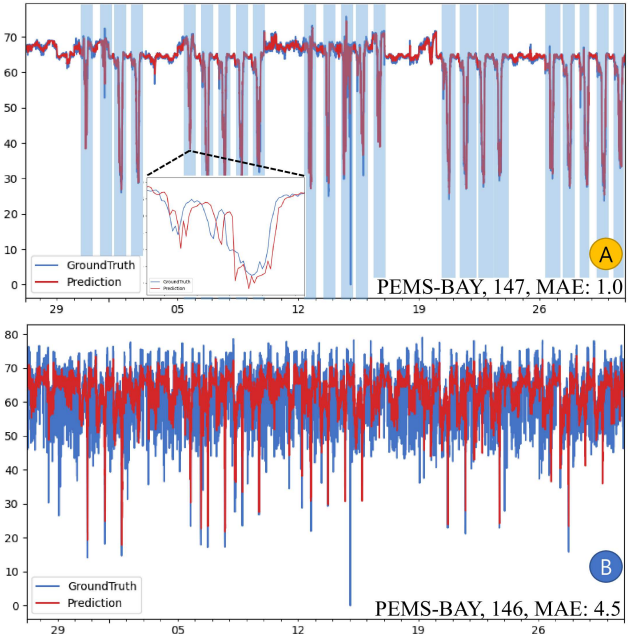}
    \caption{Different models show varying accuracy performance based on traffic data dynamics (e.g., Graph-WaveNet, data: PeMS-BAY, model prediction: red line, A: MAE: 1 (Road 147), B: MAE: 4.5 (Road 146). Blue shadow areas at A indicates the intervals that have upper 25\% standard deviation.}
    \label{fig_linechart}
    \vspace{-4mm}
\end{figure}
The opposite case is also possible, where a model does not perform well on average, but its performance improves significantly when road conditions change abruptly. 
Fig.~\ref{fig_linechart} shows two example cases that describe how model performance changes according to the road's traffic condition. 
The model effectively follows the road speed trend (MAE: 1) for road 147, while the same model produces 4.5 times lower accuracy (MAE: 4.5) for road 146. 
To measure performance on different road condition, we first extract intervals by computing a moving standard deviation with a 30-minute window size. 
Then we choose the upper 25\% of the intervals for the experiment, which have steeper speed and flow fluctuations.
The vertical blue bars in Fig.~\ref{fig_linechart} (top) presents an example set of the extracted intervals, which consists of multiple abruptly changing speed trends.

Fig.~\ref{table:impeded} shows the experimental results with different models and the extracted difficult intervals in MAE.
We first notice that the overall performance distribution is changed. 
For example, ASTGCN shows worse performance across all datasets when evaluated with the entire testset, but its performance improved significantly with difficult intervals, comparable to other models.
We further investigate how much performance decline happens to each model, which allows us to compare models and find which one adapts better to abruptly changing conditions. We call these conditions \textit{difficult intervals}.
Fig.~\ref{table:impeded} shows the performance decline in the second row. 
First, we observe that all the models have large performance decline between 67.3\% to 180.3\%, compared to their previous performance. 
If we consider MAE, Graph-WaveNet and GMAN outperform other models. 
However, ASTGCN shows the lowest performance decline across all datasets, which means it is more robust than other models in making predictions when road conditions are changing abruptly.

In terms of both MAE and performance drop, ST-MetaNet shows almost the worst performance with the difficult intervals. 
This large performance decline is caused by its meta-learning approach, which generates weights from invariant prior knowledge, such as neighboring nodes.

\section{Findings and Lesson Learned}
In this section, we present our findings and lessons learned in this work.
First, although speed and flow are correlated~\cite{HCM10}, they do not have exactly the same tendencies.
In the first experiment (Fig.~\ref{fig:Model_Evaluation}), we observe that the performance of different models differs across the evaluation metrics and datasets. 
For example, MAPE shows the largest error values among the metrics with speed data, while RMSE has the largest error values with flow data. 
This result implies that using one or two metrics may not be sufficient when new models are evaluated. 

Although RNN-based sequence prediction models perform better, they often show a weakness in making long-term predictions due to their auto-regressive property, where errors are accumulated along with previous prediction errors. 
For example, ST-MetaNet shows a high performance drop between the 30- and 60-minute intervals, as shown in Fig.~\ref{fig:Model_Evaluation}. 
Compared to ASTGCN, ST-MetaNet perform better on the short-term predictions, but worse on the long-term predictions with most of the flow datasets. 
Thus, we can assume that RNNs' long-term prediction accuracy is highly dependent on their short-term predictions.
We extract difficult intervals with a 30-minute window and observe that all models record a declined accuracy. 
This result implies that the model performance is related to the (moving) standard deviation of intervals.
For example, we observe that when the speed changes slowly in a short window, as shown in Fig.~\ref{fig_linechart} A, the model easily predicts the future pattern. In contrast, when the speed abruptly changes, the models cannot predict changes in the pattern (Fig.~\ref{fig_linechart} B), resulting in declined accuracy. 
A future study should investigate this question further, as answering it could lead to dramatic improvements in model performance. 

\section{Conclusions and Future Work}
Although numerous deep learning models have been proposed, it has been difficult to compare them and determine which models are better for predicting road speed and flow. 
In this work, we conduct a series of experiments with the state-of-the-art models to compare average accuracy, inference time, and accuracy with difficult intervals. 
All experiments are performed in the same conditions and with the same datasets. 
The results indicate that Graph-WaveNet shows the best average performance and GMAN has an advantage in long-term predictions. 
They also show better performance than others for the difficult intervals. 
Lastly, the different performance of the models with different intervals and difficulties suggest that future research on traffic prediction models should consider evaluating models under a variety of traffic conditions for evaluation. 
This work does not answer the question on why model performance differ by traffic data patterns. 
A future study can investigate to answer the question and seek to reduce the performance gap among different traffic patterns.

\section*{Acknowledgment}
This work was supported by Institute of Information \& communications Technology Planning \& Evaluation(IITP) grant funded by the Korea government (MSIT)--No.20200013360011001, Artificial Intelligence Graduate School Program (UNIST) and the 2021 Research Fund (1.210036.01) of UNIST. This work was also partly supported by NAVER Corp.


\end{document}